\documentclass[12pt]{article}
\usepackage[T1]{fontenc}
\usepackage[utf8]{inputenc}
\usepackage{authblk}

    \usepackage{amsfonts}
    \usepackage{amssymb}
    \usepackage{amsbsy}
    \usepackage{amsmath}
    \usepackage{amsthm}
    \usepackage{bbm}
    \usepackage{bm}
    \usepackage{mathtools}

\newcounter{subassumption}[assum]
	
	\makeatletter
	\renewcommand{\p@subassumption}{\theassum}% Counter prefix.
	\makeatother

	\newtheorem*{obs*}{Observation}

    \usepackage[flushleft,online,para]{threeparttable}
	\usepackage{graphics}
\usepackage[vlined,linesnumbered,ruled,resetcount]{algorithm2e}
\SetKwInOut{Initialization}{Initialization}

    \usepackage{subfig}
    \graphicspath{{Graphs/}}

    \usepackage{tikz}
    \usetikzlibrary{arrows}
    \usetikzlibrary{decorations.pathreplacing}
    \usepackage{pgfplots}
\usepackage{indentfirst}             
\usepackage{floatrow}
    \usepackage{tabularx}
        \newcolumntype{Z}{>{\centering\arraybackslash}X}
        \newcolumntype{L}{>{\raggedright\arraybackslash}X}
    \usepackage{dcolumn}
        \newcolumntype{d}[1]{D{.}{.}{#1}}
    \usepackage{rotating}                     
    \usepackage{lscape}
    \usepackage{pdflscape}   
\usepackage{authblk}

	\usepackage{csquotes}
	\usepackage[round]{natbib}
	\usepackage{url}

    \usepackage{xcolor}
        \definecolor{darkblue}{rgb}{0,0,0.4}
    \usepackage{hyperref}
        \hypersetup{
            colorlinks = true,
            linkcolor = darkblue,
            citecolor = darkblue,
            pdfborder = 0 0 0,
            pdfdisplaydoctitle = true,
            pdfhighlight = /N,
            pdfpagelayout = OneColumn,
            pdfpagemode = UseNone,
            pdfstartview = {FitH},
            pdfauthor = {{AB, CF \& JR}},
            pdftitle = {{bits}},
            pdfsubject = {{}}
        }

    \usepackage[textsize=footnotesize, colorinlistoftodos, textwidth=4cm, obeyDraft]{todonotes}
    \usepackage{geometry}
    \usepackage{setspace}
    \usepackage[bottom, multiple]{footmisc}  
    \usepackage{verbatim}
    \usepackage[normalem]{ulem}     
    \usepackage{mathpazo}
    \usepackage[toc,page]{appendix}
	\usepackage{lipsum}

\usepackage{makecell}
\usepackage{threeparttable}
\usepackage{subfig}
\usepackage{graphicx}
\usepackage{booktabs}

\begin{document}

\begin{spacing}{0}

\title{Blending Advertising with Organic Content in E-Commerce: A Virtual Bids Optimization Approach\thanks{Please
contact the authors at \texttt{carlos.carrion@jd.com} (Carrion); \texttt{wangzenan5@jd.com}
(Wang); \texttt{harikesh.nair@stanford.edu} (Nair); \texttt{xianghong.luo@jd.com}
(Luo);  \texttt{yulin.lei@jd.com} (Lei); \texttt{xiliang.lin@jd.com} (Lin); \texttt{chenwenlong17@jd.com} (Chen); \texttt{qiyu.hu@jd.com@jd.com} (Hu); \texttt{pengchangping@jd.com} (Peng); \texttt{baoyongjun@jd.com} (Bao); or \texttt{paul.yan@jd.com} (Yan) for correspondence.}}
\end{spacing}

\author{{\small Carlos Carrion \qquad Zenan Wang \qquad Harikesh Nair \qquad Xianghong Luo \qquad Yulin Lei \qquad Xiliang Lin \qquad Wenlong Chen \qquad Qiyu Hu \qquad Changping Peng \qquad Yongjun Bao \qquad Weipeng Yan}}

\date{{\footnotesize This draft: \today}}
\maketitle
\vspace{-7bp}

\begin{abstract}
\footnotesize
\begin{singlespace}
\noindent In e-commerce platforms, sponsored and non-sponsored content are jointly displayed to users and both may interactively influence their engagement behavior. The former content helps advertisers achieve their marketing goals and provides a stream of ad revenue to the platform. The latter content contributes to users' engagement with the platform, which is key to its long-term health. A burning issue for e-commerce platform design is how to blend advertising with content in a way that respects these interactions and balances these multiple business objectives. This paper describes a system developed for this purpose in the context of blending personalized sponsored content with non-sponsored content on the product detail pages of \emph{JD.COM}, an e-commerce company. This system has three key features: (1) Optimization of multiple competing business objectives through a new virtual bids approach and the expressiveness of the latent, implicit valuation of the platform for the multiple objectives via these virtual bids. (2) Modeling of users' click behavior as a function of their characteristics, the individual characteristics of each sponsored content and the influence exerted by other sponsored and non-sponsored content displayed alongside through a deep learning approach; (3) Consideration of externalities in the allocation of ads, thereby making it directly compatible with a Vickrey-Clarke-Groves (VCG) auction scheme for the computation of payments in the presence of these externalities. The system is currently deployed and serving all traffic through \emph{JD.COM}'s mobile application. Experiments demonstrating the performance and advantages of the system are presented. 
\end{singlespace}

\begin{singlespace}
\end{singlespace}
%\noindent \vspace{7bp}
%\vspace{7bp}
%\vspace{7bp}
%\vspace{7bp}
\end{abstract}

\section{Introduction}

The importance of advertising for e-commerce companies has continued to increase steadily in the last few years. It has become increasingly common that sponsored content (i.e., paid ads) and non-sponsored content (i.e., curated material selected and displayed to users free of charge to brands; hereafter referred to as organic content) are jointly displayed to users. An example is the \emph{recommended products} section commonly displayed in the detail pages of products. Generally, recommending relevant content to a user leads to a trade-off between organic content and ad content. The typical premise is the former content is more engaging and relevant to the interests of the users compared to the latter. In addition, the former content contributes to the long-term growth and retention of users in the platform, while the latter contributes to the monetization of content, enabling a healthy ad revenue business to the platform, while helping brands to expose their products to desired target audiences. Thus, the development of a system for the allocation of ad content must consider the presence of organic content in order to avoid duplicated content, adverse competition between ads and organic content, and other possible interactions. 

Furthermore, a key conundrum for the optimal allocation of ads is satisfying multiple competing objectives (e.g. click-through rate for ads and/or organics, ad revenue, time spent in pages of certain advertised product categories, etc.). The multiple objectives may arise as representative proxies of various components of long-term platform health; or due to the fact that such a complex system usually involves many teams with different focuses in midsize to large companies. The multiple objectives may be competing with each other when improving one degrades another, leading to complex trade-offs.

\begin{figure}[h]
\begin{minipage}{.24\linewidth}
\centering
\subfloat{\label{fig:mix_instacart}\begin{frame}{\includegraphics[width=\textwidth]{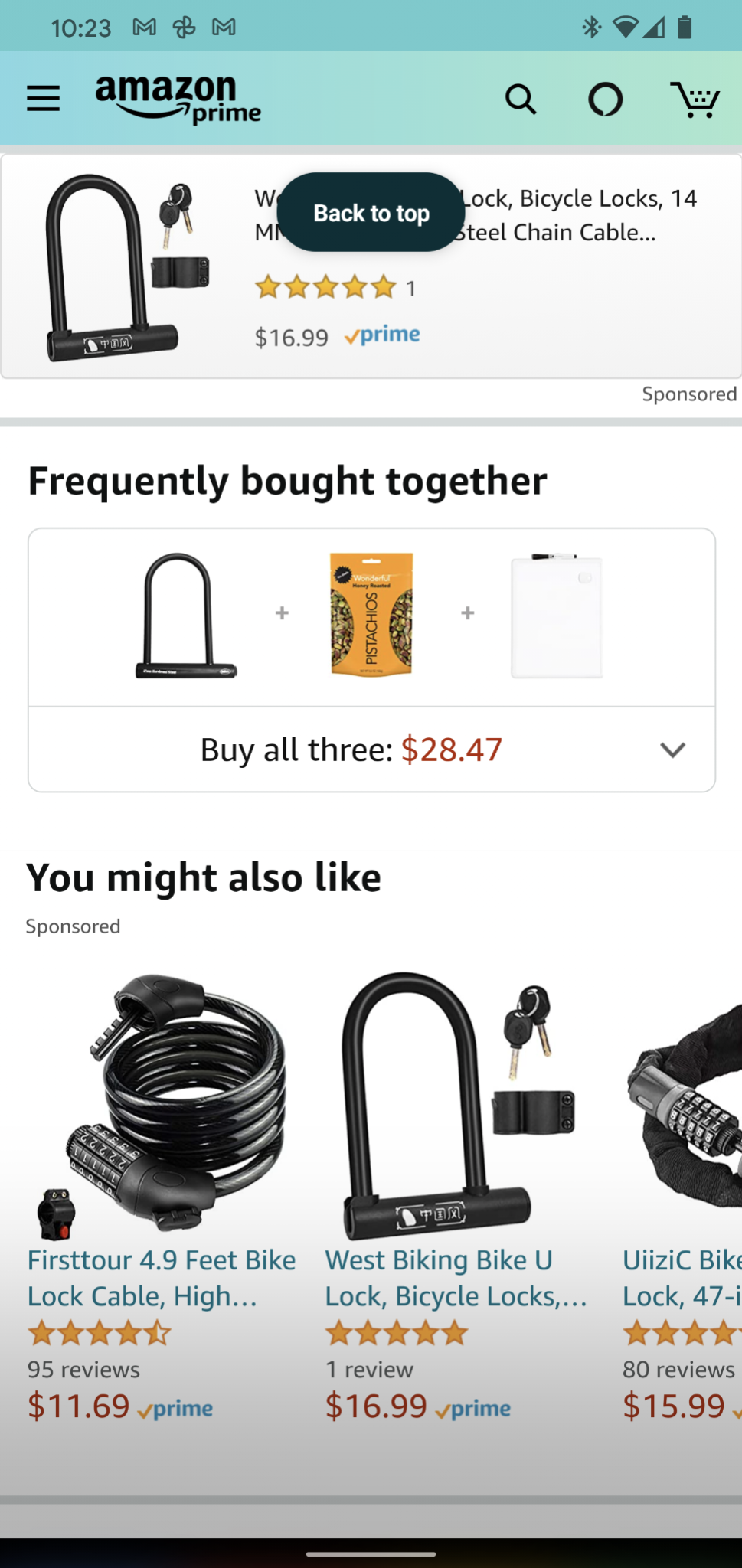}}
\end{frame} }
\end{minipage}
\begin{minipage}{.24\linewidth}
\centering
\subfloat{\label{fig:mix_Etsy}\begin{frame}{\includegraphics[width=\textwidth]{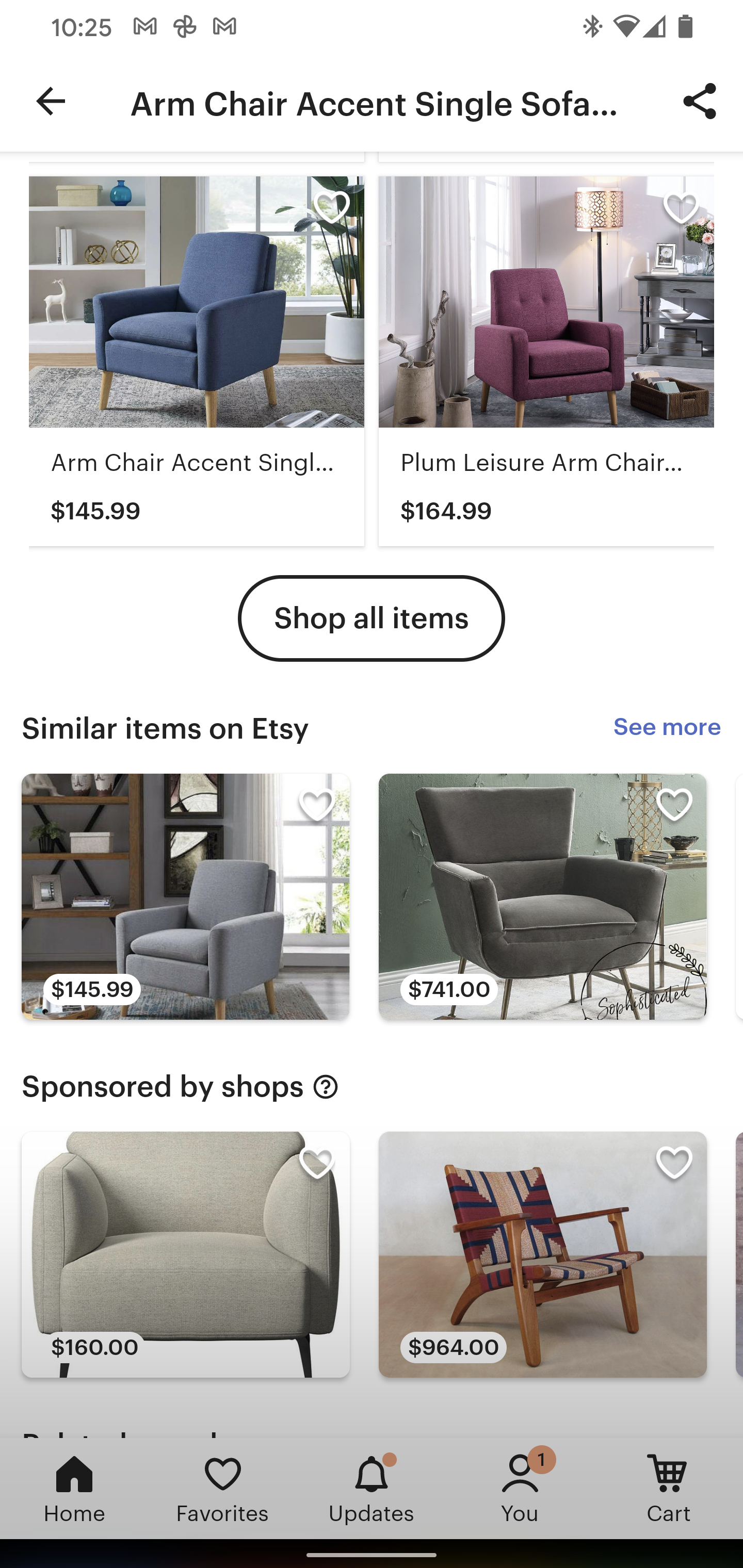}}
\end{frame}}
\end{minipage}                                                                                               
\begin{minipage}{.24\linewidth}
\centering
\subfloat{\label{fig:mix_amazon}\begin{frame}{\includegraphics[width=\textwidth]{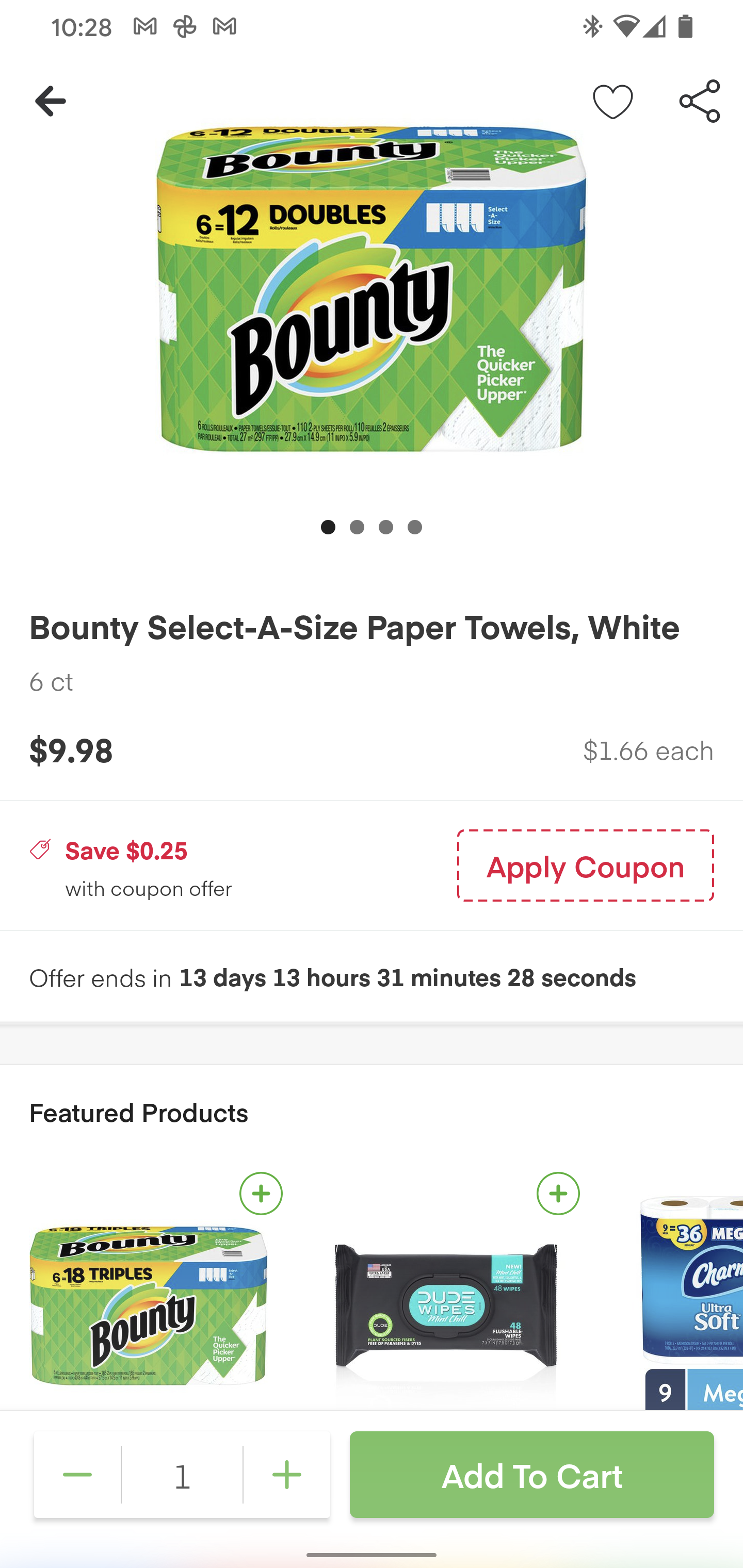}}
\end{frame}}
\end{minipage}
\begin{minipage}{.235\linewidth}
\centering
\subfloat{\label{fig:mix_JD}\begin{frame}{\includegraphics[width=\textwidth]{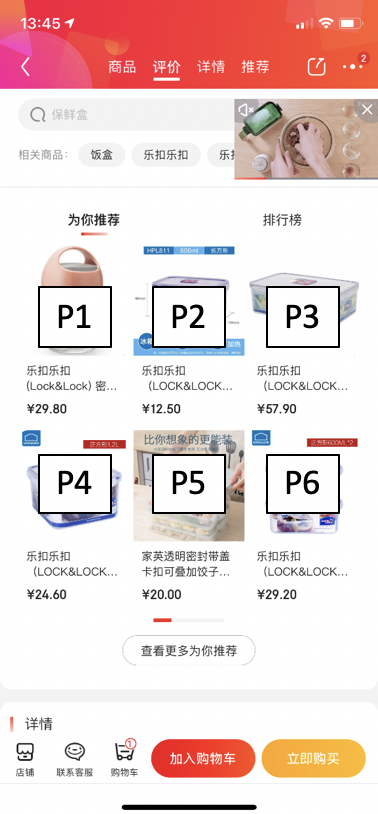}}
\end{frame}}
\end{minipage}\par  
\caption{Ads and org. recommendations on product-detail pages on E-Comm apps: (L-R) Amazon, Etsy, Instacart, JD.}
\label{fig:mixer-screenshots}
\end{figure}

In this research, we describe the development of a system for the personalized allocation of ad content in the presence of organic content and multiple objectives. Our system is applied to the product detail pages in \emph{JD.COM}'s mobile e-commerce application. A concrete example of this environment is the \emph{product recommendation} section of the page presented in Figure~\ref{fig:mixer-screenshots} for several e-commerce vendors. In the example from \emph{JD} shown in the last panel, there are 6 positions available to display recommended ad or organic content, labeled ``P1-P6''. On \emph{JD}'s mobile app, the product recommendation section is located on the product page between a panel containing a short summary of the product and user reviews, and a panel containing a detailed description of the product. Our system focuses on personalized ad allocation on these positions. All users arriving at hundreds of millions of product detail pages available on the app can potentially be exposed to such ads.

Our system has three distinctive features. Firstly, it allows analysts to flexibly accommodate the multiple competing objectives in a virtual-bid formulation to be discussed subsequently. It also offers an explicit way to balance these multiple objectives using virtual bids to find a reasonable \emph{tipping point} for the competing objectives  (i.e., improving all objectives to a point where improving one worsens another). Moreover, it
presents a novel approach for the computation of these virtual bids without requiring explicit elicitation of these bids or even the specification of constraints from stakeholders. This new approach has yet to be presented in the literature of multiobjective optimization for personalized ad allocation systems \cite{agarwal2016}. In addition, the virtual bids have an economic interpretation; they represent the implicit valuation of the platform for each of these multiple objectives, i.e., the willingness to pay. Thus, these virtual bids allow the platform to express its desired trade-off between these objectives as its implicit valuations for each of these competing objectives of interest relative to the ad revenue, e.g., marginal rates of substitution between ad revenue and user engagement for ads. This aids interpretability and may have independent business decision value for other related questions (such as deciding optimal ad load on the platform). A second feature is the explicit modeling of users' click behavior as a function of their characteristics, the individual characteristics of each sponsored content displayed, and more importantly, the influence exerted by other ad and organic content displayed alongside. Thus, our system offers a unified modeling solution that allows the platform to implement personalized ad allocation while capturing flexibly relevant \emph{joint effects} (i.e., any substitution and/or complementarity effects within ads and between ads and organics).

In our system, the \emph{joint effects} and the \emph{virtual-bid approach} induce an allocation mechanism requiring a payment mechanism that properly constructs charges recognizing the \emph{externalities} imposed by an arbitrary ad content on other ad and organic content. Thus, the third feature of our allocation of ads mechanism is its compatibility with the VCG auction mechanism \cite{tim2016}. Generally, the widely used \emph{Generalized Second Price} (hereafter referred to as GSP) auction mechanism does not consider such externalities in the payment mechanism \cite{gomes2009externalities, roughgarden2012externalities}.

The proposed system has been deployed and it serves on average tens of millions of auctions per day. In the rest of this paper, we describe first our proposed approach (Section \ref{sec:framework}); and report on a battery of experiments to evaluate the performance of the proposed approach (Section \ref{sec:experiment}). Section \ref{sec:relwork} reviews the literature to contrast our proposed approach with related work and Section \ref{sec:conclusion} concludes. \\

\section{Application Set-up}\label{sec:framework}
The applied problem we tackle is to decide how to do personalized allocation of ads to users arriving at a product detail page. The allocation needs to respect possible interactions with others ads and organics on the page, so as to optimize against multiple, possibly competing, platform objectives. While our procedure can be framed and implemented more generally, we present it in the context of some specific decisions made in its deployment at \emph{JD.com}. First, motivated by considerations of latency and separation of ad and organic product development, we focus only on ad allocation, and take the allocation of organic content on the page as given. \cite{yan2020} provides an excellent discussion of various reasons why such ``de-coupling'' is desirable and common in large organizations. Second, while conceptually straightforward, due to organizational reasons, both the number and the location of ad slots on the page were pre-determined and not optimized in this deployment. Hence, we optimize only the identity and the ordering of ads, and do not address the problem of optimal ad load or dynamic ad location. Third, we present the discussion in the context of two objectives to be balanced in the allocation; it is straightforward to allow for more objectives. The objectives discussed in this application were pre-determined by platform management and are taken as given in the development of the system.
\subsection{Key Challenges}\label{sec:application-setting}
For a user arriving at a product's detail page, we need to select $K_{ads}$ to be placed in $P_{ads}$ fixed positions given $K_{orgs}$ organics already placed in $P_{orgs}$ fixed positions. The fixed positions are decided a priori from positions P1-P6 on the \emph{product recommendation} section in Figure~\ref{fig:mixer-screenshots}. The cardinality of $P_{ads}$ is $K_{ads}$. The $K_{ads}$ selected for display must be optimal with respect to two objectives: (1) Expected ad revenue obtained from a Click-per-Cost (hereafter referred to as CPC) billing mode, where the CPC charge is determined from an auction mechanism; and (2) Expected CTR from the displayed ad content. This second objective is a metric representing the user engagement with the ad content. The two objectives may not be aligned due to joint effects as explained below. There are two key challenges in optimal ad allocation:

\begin{enumerate}
    \item \emph{Externalities}. There are generally two types of joint effects within ads and across ads and organics that may influence users' click behavior: (1) identities of the ads and organics; (2) arrangement of the ads and organics. The former refers to the possible substitution and complementarity effects within the ads and also between ads and organics. For example, displaying jointly multiple ads for pencils of different colors can induce substitution within the ads, and/or complementarity between the ads and organics, when organics are erasers. The latter refers to effects produced by displaying different permutations with the same ads and organics. In many ad-contexts, for e.g., in traditional search advertising, ads and organics are selected and ranked separately, neglecting such joint effects between such content, so that the externalities arising in the allocation scheme are ignored, partly because it makes the resulting auction more complex \cite{gomes2009externalities, roughgarden2012externalities}.
    \textbf{}
    \item \emph{Dual Objectives}.  For the platform, choosing the optimal set of ads $K_{ads}$ given a set of organics $K_{orgs}$ is a balancing act between maximizing ad revenue and user engagement. An approach followed in the literature (i.e. multiobjective optimization of ads in recommendation systems \cite{agarwal2012personalized, agarwal2016,yan2020}) is to frame the problem as maximizing an unconstrained platform objective by assigning \emph{shadow prices} to the ad-CTRs (i.e. the user engagement for ads in our case) in order to evaluate both the engagement objective and the ad revenue objective in the same units (i.e., currency). However, an open challenge is to properly choose the shadow prices. Alternatively, the problem has been formulated in this literature \cite{agarwal2012personalized, yan2020} as a constrained optimization problem where the shadow prices are traded for minimum thresholds for the user engagement goal, CTR for e.g., and then optimal ad allocation is found by maximizing ad revenue subject to this constraint. The challenge now is the minimum thresholds must now be properly set. A strategy generally followed in this literature is to use a fraction of the optimal CTR obtained by solving an unconstrained maximum CTR problem as the minimum thresholds, but this creates a new hyper-parameter to set, which is the fraction. Therefore generally, a significant challenge in both approaches is the need to determine a set of critical hyper-parameters representing appropriately the trade-off between the objectives the platform. \textbf{The principled determination of these hyperparameters, which is a major focus on this article, is barely discussed in the literature \cite{agarwal2016}}. Further, another source of complexity is when value of an ad slot to an advertiser depends on who else is around her, there is no natural ordering of slots. As a consequence, the standard GSP auction's payment scheme cannot be directly applied \cite{gomes2009externalities, roughgarden2012externalities}.
\end{enumerate}
\subsection{Existing System for Ads Allocation}
The allocation system in place prior to the deployment of the new system consists of three key components: (1) a deep learning based model (hereafter referred to as the \emph{pointwise} model) for predicting the CTR of an ad as a function of the characteristics of the user and the characteristics of only that ad considered separately from others and organics (for exemplar architecture, see \cite{covington2016deep}); (2) an allocation algorithm that ranks ads by computing an exponentially weighted \emph{effective cost per mille} (hereafter referred as ``eCPM'') for each ad, with a hyperparameter $t$ tuned using experimentation, e.g., the weighted eCPM for an ad is $bid_{CPC} \times pCTR^{t}$ where $bid_{CPC}$ is the (CPC) bid submitted by the advertiser and $pCTR$ is the predicted CTR of the ad; and (3) a GSP auction mechanism to compute the corresponding payment given the ads allocation. We refer to this as the baseline system and use it as a benchmark in our experiments later.
%A limitation of this deployed system is neglecting the presence of the organic content in the allocation of ads in addition to the challenges discussed previously about \emph{externalities} and \emph{dual objectives of ad revenue and user engagement for ads}. %In addition, it must be noted that the hyper-parameter $t$ attempts to balance both objectives although this approach lacks transparency and it is devoid of economic meaning in contrast to the virtual bids approach.

\subsection{Proposed New System for Ads Allocation}

The proposed new system consists of three key components in addition to the components of the baseline system: (1) An enhanced deep learning model (hereafter referred to as the \emph{listwise} model) for predicting the CTR of ads considering the joint effects from other advertisements and organics displayed together; (2) a listwise ranking of ads using a \emph{virtual bids} composite objective function; and (3) a listwise generator to enumerate multiple lists of size $6$ of ads and organic content. 

\begin{figure}[h]
  \centering
  \includegraphics[width=0.85\linewidth]{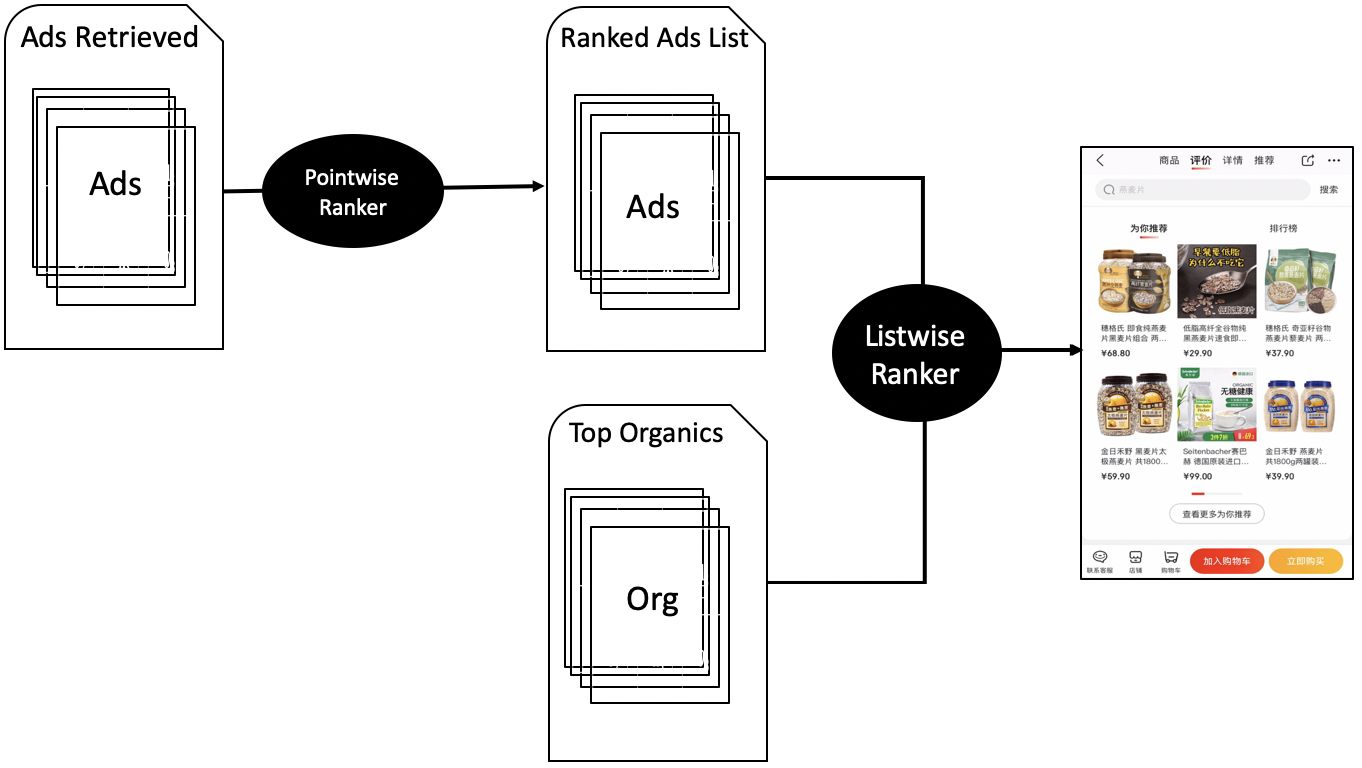}
  \caption{System overview in for an arbitrary product's detail page in the JD.COM mobile app}
  \label{fig:system}
\end{figure}

\begin{figure}[h]
  \centering
  \includegraphics[width=0.4\linewidth]{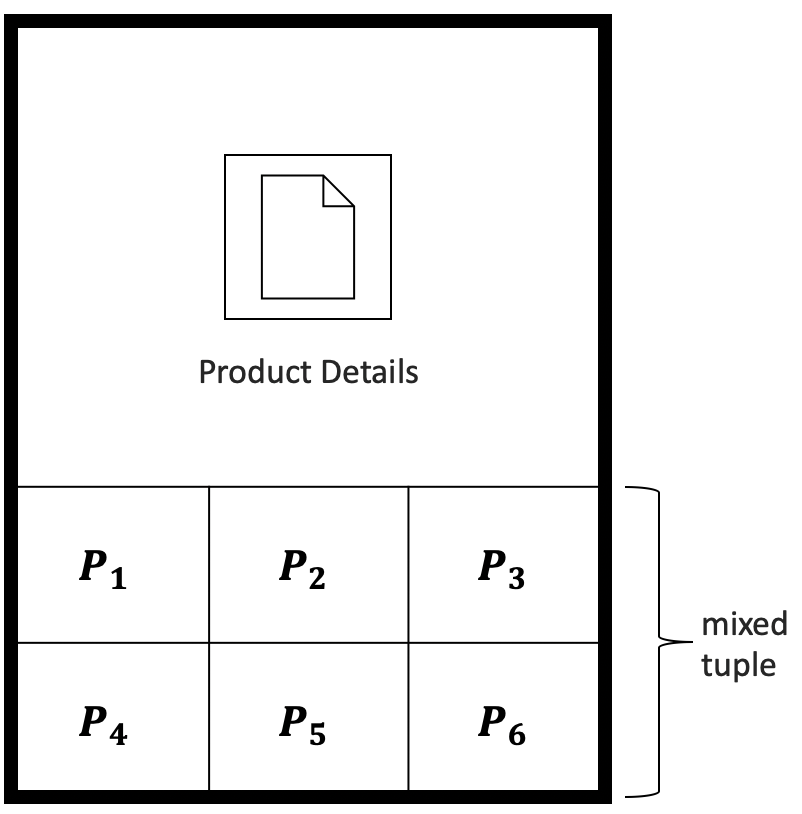}
  \caption{Graphical representation of the unit of analysis for the proposed system for ads allocation}
  \label{fig:mixer}
\end{figure}

The unit of analysis in the proposed system is a \emph{mixed tuple}, denoted $\omega$. $\omega$ is of size $6$ composed of $K_{ads}$ ads and $K_{orgs}$ organics (i.e. $K_{ads} + K_{orgs} = 6$) placed in $6$ positions P1-P6 as shown in Figure~\ref{fig:mixer}. For each arriving user, the proposed system takes as inputs, a pre-ranked list of candidate ads $A_{1:N_{ads}}$ of size $N_{ads}$, and a ranked list of organic content $O_{1:N_{orgs}}$ of size $N_{orgs}$. The list of candidate ads is pre-ranked using the baseline system; and the organic list is ranked based on a model outside of ads. Figure~\ref{fig:system} presents a schematic.  

\noindent For an arriving user, a candidate $\omega$ is generated by choosing the top $K_{orgs} < 6$ organics of the ranked list $O_{1:N_{orgs}}$ and $K_{ads}=6 - K_{orgs}$ ads from the ranked list $A_{1:N_{ads}}$. The set of possible $\omega-$s is denoted $\Omega$, whose size is $K_{ads}! \times \binom{N_{ads}}{K_{ads}}$ because the order of ad placement matters. Since there is a set $P_{orgs}$ of positions for organics that is fixed, and the placement of the top $K_{orgs}$ organics from $O_{1:N_{orgs}}$ in this set are decided a priori, the organic components of all candidate $\omega-$s considered will be the same. However, the identity and order of the $K_{ads}$ placed in $P_{ads}$ will vary across candidate $\omega-$s, and is what we optimze over. 

\subsubsection{Optimization Problem}
We let $x(\omega)$ denote the $K_{ads} + K_{orgs}$ dimensional vector of predicted CTRs for the components of $\omega$. The fact that $x(\cdot)$ is indexed by $\omega$ reflects the joint effect of the ads and organic components on each other. Denote the $K_{orgs}$ organic components in $\omega$ as $\omega^o$, and the $K_{ads}$ ads as $\omega^a$. Then, we denote the predicted CTR of the $j$th organic element $\omega_{j}^{o}$ of $\omega^{o}$ as $x_{j}^{o}(\omega)$; and the predicted CTR of the $j$th ad element $\omega^{a}_{j}$ of $\omega^{a}$ as $x_{j}^{a}(\omega)$.

We find the best allocation by searching for the $\omega$ that solves the following problem,
\begin{align}
\max_{I_{\omega}:\omega\in\Omega}\sum_{\omega\in\Omega}^{|\Omega|}\sum_{j=1}^{K_{ads}}v^{a}x_{j}^{a}(\omega)I_{\omega}+\sum_{\omega\in\Omega}^{|\Omega|}\sum_{j=1}^{K_{ads}}b_{j}x_{j}^{a}(\omega)I_{\omega}\enskip s.t.\sum_{\omega\in\Omega}^{|\Omega|}I_{\omega}\leq1\label{eq:virtual_bid}\end{align}

\noindent The term to be maximized is a composite objective, reflecting management's stated interest in serving ad allocations that respect both the expected ad-CTR and the expected ad revenue. Eq. \ref{eq:virtual_bid} converts all sub-objectives to the same units (money), so that the overall objective is additive. The first term, which we refer to as money-metric ad-CTR uses a parameter $v^{a}$ to convert expected ad-CTR into money metric terms; while the second term representing ad revenue is already in monetary terms. The control variables $I_{\omega} \in \{0, 1\} \forall \omega \in \Omega$ are binary, and therefore (\ref{eq:virtual_bid}) is an integer linear program where we search for the optimal $\omega^{*}$ by setting $I_{\omega'} = 1$ if $\omega^{*} = \omega'$ for the solution that maximizes the composite objective function.

We call the formulation in Eq.~\ref{eq:virtual_bid} the \emph{virtual-bid problem}, because the virtual bid $v^{a}$ can be interpreted as the platform's valuation for one click of ad content. This is analogous to the advertisers' CPC bids submitted to the platform, which economically speaking, represent the advertisers' valuations for each click on their ads. In a VCG auction context, this formulation has a nice interpretation of both the platform and the advertisers bidding jointly for the user impression. On the other hand, the challenge of this formulation is that it is difficult to elicit from the platform, its implicit valuations encapsulated in the virtual bids (i.e. willingness to pay for clicks on ads). Further discussion of this issue is deferred to Section~\ref{subsec:get_virtual_bids}, where an alternative data-driven solution is provided. 

Assume for a moment that $v^{a}$ is known, and that for each $\omega\in\Omega$, we can obtain $x(\omega)$. Algo~\ref{alg:reranker} presents pseudo-code for the implementation of a linear search to find the optimal $\omega^{*}$ for the \emph{virtual bids problem} taking $v^{a}$ and $x(\omega)$ as given. 
\begin{algorithm}[h]
\caption{Listwise Ranker}
\label{alg:reranker}
\KwIn{$L_{\omega}$ is the List of mixed tuples where each mixed tuple has $K_{ads}$ ads and $K_{org}$ organics, $H$ Map of CPC Bids for ads, and the virtual bid $v^{a}$}
\KwOut{$L^{*}$: List of Optimal mixed tuple}
Initialize $L^{*} = []$, Obj$_{max} = 0$ \\
\For{$\omega \leftarrow  L_{\omega}$} {
$i = 0$ \\
Obj$_{\omega} = 0$ \\
\For{$i \leftarrow \omega$.getAdsIndices()} {
Obj$_{\omega} += \omega(i)$.getPCTR() * ($v^{a}$ + H($\omega(i)$.getAdID())) 
}
  \If{Obj$_{\omega}> $Obj$_{max}$} {
    Obj$_{max} = $ Obj$_{\omega}$ ;
    $L^{*} = \omega$
  }  
}
\KwRet{$L^{*}$}
\end{algorithm}
In practice, the cardinality of the set $\Omega$ is likely to grow very large, leading to concerns about latency in ad-serving. To address this, the deployed system implements a heuristic version of Algo.~\ref{alg:reranker}. We construct $\omega$ choosing only the top $N'_{ads}$ of the pre-ranked list of ads $A_{1:N_{ads}}$ where $N'_{ads} << N_{ads}$, restricting $|\Omega| = K_{ads}! \times \binom{N'_{ads}}{K_{ads}}$. This heuristic assumes that the top $N'_{ads}$ are likely to have the largest joint effects. $N'_{ads}$ is treated as a hyperparameter that is tuned from experimentation.
\subsubsection{Comparison to Framing as a Constrained Ad Revenue Problem}\label{sec:compare-constrained-approaches}
Before discussing how we obtain $v^{a}$ and $x(\omega)$, we briefly compare the approach outlined in (\ref{eq:virtual_bid}) of solving the unconstrained problem directly, to alternative approaches for multi-objective ad allocation that solve constrained versions of (\ref{eq:virtual_bid}). An alternative to solving Eq. \ref{eq:virtual_bid} is to set a minimum desired CTR threshold for ads, $C$, and find the allocation that maximizes ad revenue subject to this constraint, i.e.,
\begin{align*}
 \max_{I_{\omega}:\omega\in\Omega}\sum_{\omega\in\Omega}^{|\Omega|}\sum_{j=1}^{K_{ads}}b_{j}x_{j}^{a}(\omega)I_{\omega}\enskip s.t.\enskip\sum_{\omega\in\Omega}^{|\Omega|}I_{\omega}\leq1;\sum_{\omega\in\Omega}^{|\Omega|}\sum_{j=1}^{K_{ads}}x_{j}^{a}(\omega)I_{\omega}\geq C
\label{eq:const_ads}
\end{align*}
As mentioned in Sec. \ref{sec:application-setting}, this now requires setting the thresholds $C$ properly, which is difficult; and often requiring solving auxiliary sub-problems. If $C$ is chosen arbitrarily, it is possible that solving the constrained problem above does not deliver the ad-CTR or ad revenue that solving (\ref{eq:virtual_bid}) can deliver. More unfortunately, even with a well-chosen $C$, it is difficult to guarantee that solving the constrained problem in general will deliver the same objective as solving the unconstrained problem directly. Guaranteeing this requires \emph{strong duality} to hold, which for integer linear programs requires more restricted conditions than linear programs, see \cite{bertsimas2005, fisher1981}. From this perspective, our approach has two advantages. First, recognize we can interpret Eq. \ref{eq:virtual_bid} as a \emph{Lagrangian relaxation} of the constrained ad revenue problem above. Following standard results (e.g., \cite{fisher1981}), therefore we are guaranteed that solving Eq. \ref{eq:virtual_bid} provides an upper bound to the constrained ad revenue problem. The link to Lagrangian duality also provides additional interpretability to our virtual bids, as we can interpret them as shadow prices that penalize ad revenue in accordance to the violation of implicit ad-CTR constraints. Secondly, below, we will provide a principled approach to tune the virtual bids in the unconstrained problem, which achieves a pareto optimal trade-off across component sub-objectives while maximizing the overall compositive objective.
\subsubsection{Obtaining Virtual Bids}\label{subsec:get_virtual_bids}
If one interprets the virtual bids as the willingness to pay of the platform for clicks from users on the ads, in principle, one could consider eliciting $v^{a}$ directly from key decision-makers. However, it may be difficult for decision-makers to articulate this construct in monetary terms (an \emph{expressivity} problem). Also, the willingness to pay may be dynamic, and change based on platform competition. Further, heurtistically picking $v^{a}$ can reduce ad revenue and potentially ad-CTR (due to externalities). This motivates a principled, data-driven approach to obtaining the virtual bids. Our proposed solution to obtaining  $v^{a}$ discussed below will have the characteristic that it is pareto efficient; that is, moving from the optimal $v^{a}$ to another value, leads to a trade-off between improving one of the sub-objectives in (\ref{eq:virtual_bid}) while making worse the other.

\subsubsection{Obtaining Virtual Bids from Historical Data}
\label{subsec:vb}
To obtain $v^{a}$, we solve a separate sub-problem in historical data. Collect data from an epoch of $i=1,..,N$ past impressions, and for an impression  $i$ in the data, define $V_{ia}^{*}(v^{a})$ and $V_{ir}^{*}(v^{a})$ as the optimized values of the money-metric ad-CTR and ad revenue terms obtained by maximizing the platform's composite objective in Eq. \ref{eq:virtual_bid} holding the virtual bid fixed at $v^{a}$. We define the \emph{Utopia Point} of the sub-problem as the tuple $(\bar{V}^{u}_{a}(v^{a}),\bar{V}^{u}_{r}(0))$, where $\bar{V}^{u}_{a}(v^{a}) = \frac{1}{N} \sum_{i=1}^{N} V^{*}_{ia}(v^{a})$, and  $ \bar{V}^{u}_{r}(0) = \frac{1}{N} \sum_{i=1}^{N} V^{*}_{ir}(v^{a}=0)$, i.e., the best the platform could do on average across the $N$ impressions in terms of its objective, if it cared only about each sub-objective in isolation, ignoring the other.

\begin{figure}[h]
  \centering
  \includegraphics[width=0.8\linewidth]{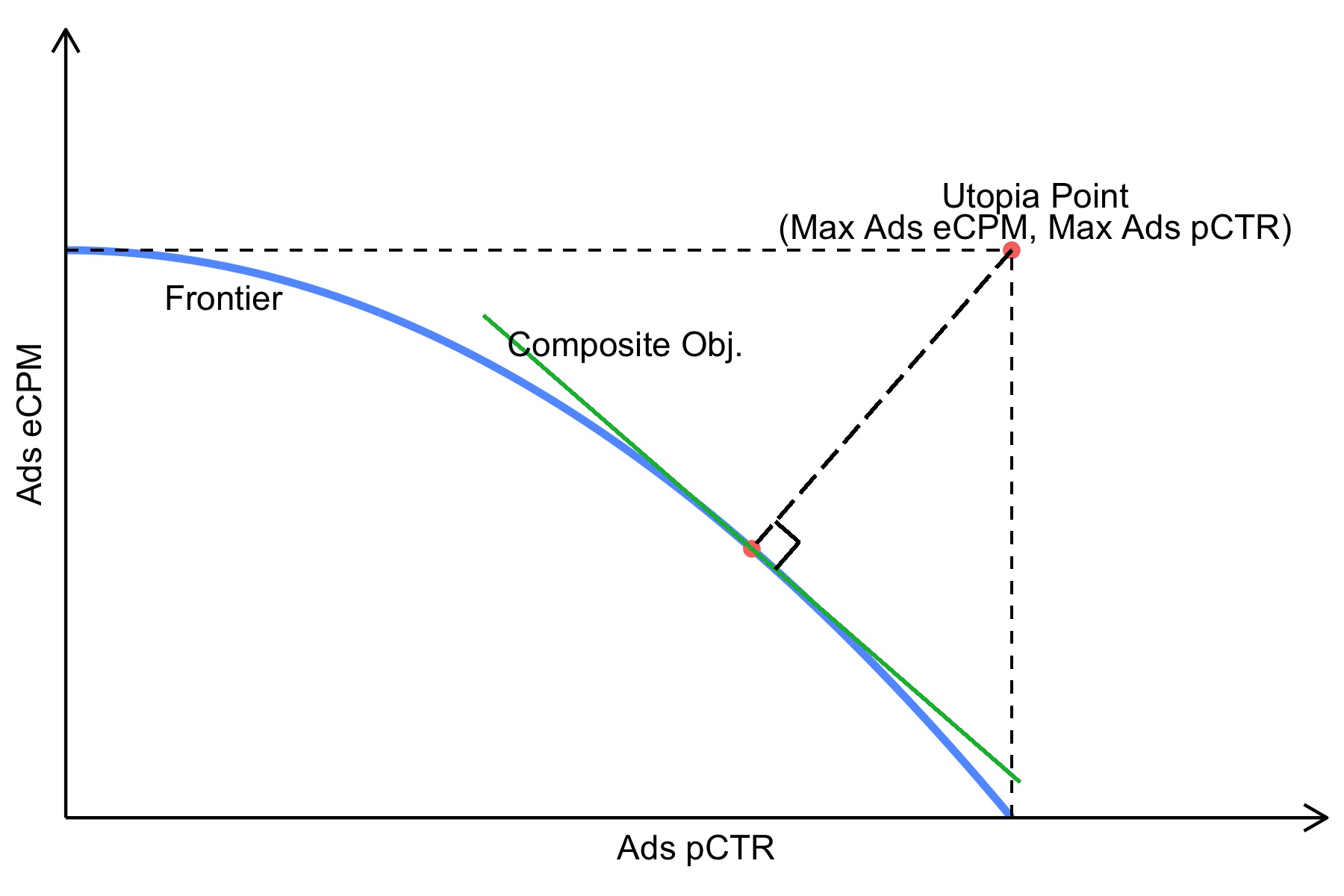}
  \caption{Stylized Plot of Bi-dimensional Possibility Frontier and Utopia Point}
  \label{fig:stylized_plot}
\end{figure}

The idea of our method is to find a $v^{a}$ so that the induced money-metric ad-CTR and ad revenue comes as close to the \emph{Utopia Point} as possible on average across the $N$ impressions. We do this by locating a point on the \emph{possibility frontier} of ad-CTR and ad revenue that is as close as possible to the \emph{Utopia Point}. The possibility frontier is the surface defined by all possible combinations of ad-CTR and ad revenue that can be generated by solving (\ref{eq:virtual_bid}) for a given vector of virtual bids that is pareto efficient. Generally the \emph{Utopia Point} is not a point on the surface of the possibility frontier of ad-CTR and ad revenue; if it was, the virtual bid should be $0$ by definition as there is no trade-off between these two objectives. See Fig. \ref{fig:stylized_plot} for graphical intuition. 

We implement this as a search for $v^{a}$ in a bilevel optimization problem. In the upper level, we search for a $v^{a}$ that minimizes the \emph{L2-norm} distance between the \emph{Utopia Point} and the tuple of averages across all $N$ impressions in the data of optimized money-metric ad-CTR and ad revenue, denoted $(\bar{V}_{a}^{*}(v^{a}),\bar{V}_{r}^{*}(v^{a}))$. $(\bar{V}_{a}^{*}(v^{a}),\bar{V}_{r}^{*}(v^{a}))$ are obtained in turn, by solving a nested, lower-level problem for each $v^{a}$ considered by the upper level. In particular, for each candidate $v^{a}$ from the upper level, the lower level solves the \emph{virtual bids problem} in Eq.~\ref{eq:virtual_bid} for each $i$, and computes the averages across $i$ of the tuple of optimized ad-CTR and ad revenue. Formally, we solve,
\begin{equation}
\begin{aligned}\min_{v^{a}} & \left[(\frac{\bar{V}_{a}^{*}(v^{a})}{\bar{V}_{a}^{u}}-1)^{2}+(\frac{\bar{V}_{r}^{*}(v^{a})}{\bar{V}_{r}^{u}}-1)^{2}\right]^{\frac{1}{2}}\\
\textrm{\enskip s.t.}\enskip & \bar{V}_{a}^{*}(v^{a})=\frac{1}{N}\sum_{i=1}^{N}V_{ia}^{*}(v^{a});\enskip\bar{V}_{r}^{*}(v^{a})=\frac{1}{N}\sum_{i=1}^{N}V_{ir}^{*}(v^{a})\label{eq:bilevelopt}
\end{aligned}
\end{equation}
\noindent For the case of one virtual bid $v^{a}$ (our current application), the upper level is one dimensional; we can solve this program efficiently using the \emph{Golden search method} (see \cite{luenberger2010}, pp. 216-219). For the case of more than one virtual bid (for example if organics CTR was another objective with virtual bid $v^{o}$), we can solve the upper level using \emph{Simultaneous Perturbation Stochastic Approximation} (SPSA) \cite{spall1998}. For completeness, these methods are briefly outlined in Appendix \ref{app:optim-algos}. In both cases, the lower level problem is solved impression by impression using Algo. \ref{alg:reranker}. 

While Program \ref{eq:bilevelopt} is solved offline, it can be updated frequently with recent logged data so the computed virtual bids reflect relevant changes in the platform environment. The frequency of updates will be platform specific, depending on the nature of the CTRs, the information content of the data, and execution time. Our experiments below document that re-tuning  $v^{a}$ in this manner is important for effective performance (which again motivates the need for a principled approach to do so). In addition, although in this implementation we compute one value of $v^{a}$ per epoch for use in ad-serving, conceptually, it is straightforward to allow it to vary by product-category, time-of-day, and other contexts, by computing Program \ref{eq:bilevelopt} separately for data subseted by these variables; or by pooling the data and making $v^{a}$ a function of these variables. This may allow the method to achieve further improvements than the lifts documented in the experiments below. %Lastly, the optimal virtual bids obtained can be further fine tuned with experimentation.

\subsubsection{Predicting CTRs for a mixed tuple}
\begin{figure}[h]
  \centering
  \includegraphics[width=0.85\linewidth]{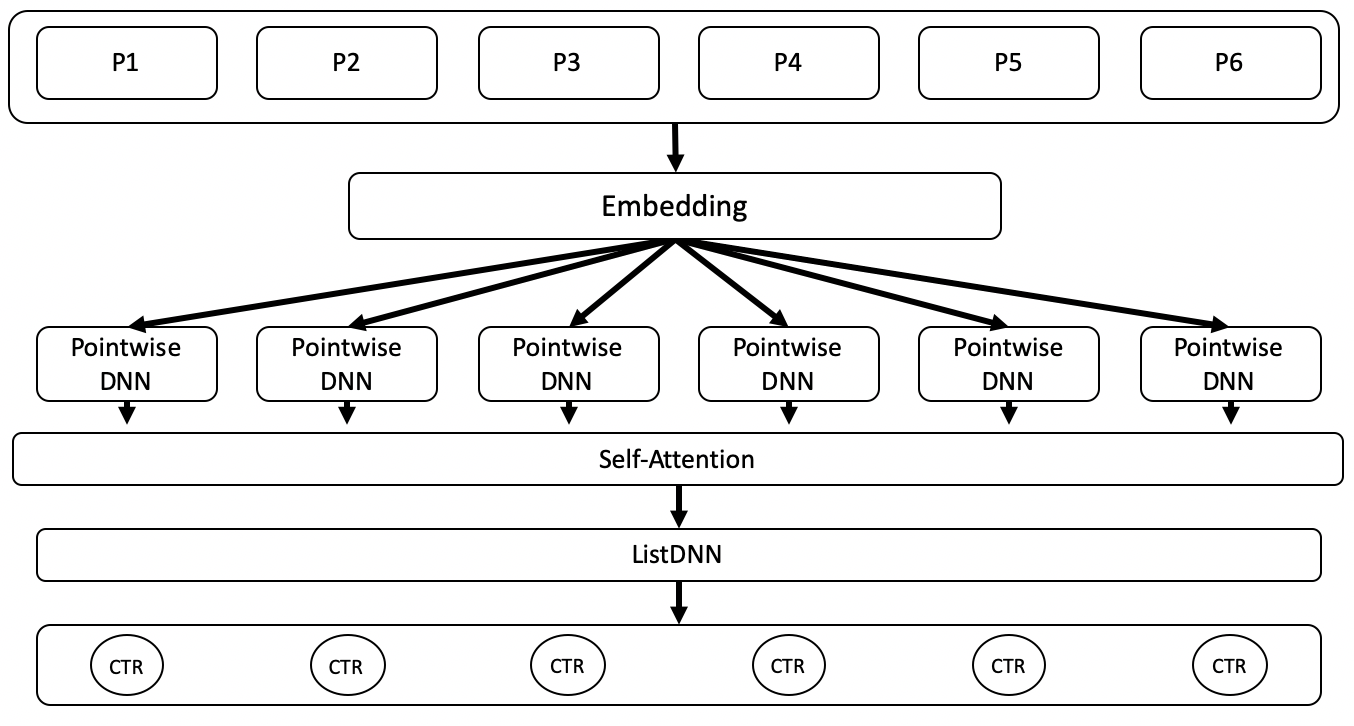}
  \caption{Architecture of listwise model for CTR prediction.}
  \label{fig:dnn}
\end{figure}

To finish the description of the system, we also need a way to obtain the predicted CTRs, $x(\omega)$ for each candidate $\omega$. We do this using the listwise model. The novelty of this enhanced deep learning model here is not as much in terms of its  architecture and training, but its use in conjunction with the problem solutions in (\ref{eq:virtual_bid}) and (\ref{eq:bilevelopt}) for ad allocation with externalities; hence, the description is intentionally brief (for a more in-depth discussion of such models, see for instance \cite{pei2019personalized}). The model uses as input features user; ads/organics; and contextual characteristics. The first and second are self-explanatory and the third category corresponds to features describing the product in the product detail page. Figure~\ref{fig:dnn} presents the model architecture. Briefly, (1) Categorical features are fed into embedding layers; (2) Numerical features are fed into Fully Connected Layers along with the Embedding Layers for the categorical features; (3) Sigmoid output layers are used for the prediction. Six dense networks are used to learn product low-dimensional representations from features. The Output Layer has six Sigmoid units corresponding to the mixed tuple; and the self-attention layer utilizes a multi-head self-attention mechanism to learn interactive and context information among six products. The model is trained with an entropy loss function and batch normalization is used. Training is done via an offline-online scheme, i.e., training data is logged from online experiments where traffic is randomly allocated to the two models, and also retrained is done often. The training set for the listwise model consists of the set of features and the whole observed mixed tuple with realized clicks on any of the six positions in contrast the pointwise model which only sees the position with realized clicks.  The model is large-scale as in typical ad-industry applications, with embedding size of about 500 million and about 4 billion parameters. In offline validation, the listwise model produces a lift of $2\%$ in the AUC of CTR prediction compared to the pointwise model.

%Furthermore both models are trained with the entropy loss function and batch normalization is used, see \cite{goodfellow2016}. In addition, the training set for the listwise model consists of the set of features and the whole observed mixed tuple with realized clicks on any of the six positions in contrast the pointwise model only sees the a position with realized clicks.

%The Pointwise and Listwise models are trained following an explore-exploit scheme (\cite{agarwal2016}) scheme, i.e. a randomly selected fraction of the users are shown a randomly selected ad for the case of the pointwise model, and a randomly selected \emph{mixed tuple} for the case of the listwise model. In addition, the training set for the listwise model consists of the set of features and the whole observed \emph{mixed tuple} with realized clicks on any of the six positions in contrast the pointwise model only sees the a position with realized clicks. Moreover, both models are trained with the entropy loss function, and batch normalization is used, see \cite{goodfellow2016}. In the appendix, we provide more details about our deep learning models.

\section{Experiments}\label{sec:experiment}
This section presents experiments run on \emph{JD.com}'s mobile app to assess the effectiveness of our proposed approach on a variety of scenarios and metrics. In addition, we report experiments to demonstrate the existence of \emph{joint effects} in our application, and study how accommodating shifts in the environment is important for effectively choosing virtual bids. In all experiments, impressions on product-detail pages are randomized into treatments, wherein a treatment is defined as a particular way of doing ad allocation. When reporting results, we use the lift transformation for all the metrics in order to protect proprietary information. Except in the last experiment in Section \ref{subsec:dist_shift}, the lifts reported in all the experiments are obtained by comparing a model $A$ against the baseline system $B$. Specifically, the lift transformation is  \texttt{lift}$_{A} = 100 \times \frac{\texttt{metric}_{A} - \texttt{metric}_{B}}{\texttt{metric}_{B}}$.
Further, the statistical significance of the results in the subsequent tables follow the convention: * for $1\%$ significance, and ** for $0.1\%$ significance. For the figures, the confidence intervals are for $99\%$ confidence. Also, in some figures, the $x$-axis or $y$-axis are deliberately removed to protect proprietary information. Lastly, the ads payment scheme used in these experiments remains the GSP payment scheme. In particular, the GSP payments are computed using the ranking of ads by weighted eCPM using the pointwise model's predicted CTR but using the ads allocation from the listwise ranker.

\subsection{Existence of Joint Effects}
There are three treatments in this experiment: Random \emph{shuffle} of the order of optimal ads from the baseline model; Random selection of ads from the top $X$ ($X=K_{ads}+1$ and $X=K_{ads}+2$) of the pre-ranked ads list and placement of these in random order on the available ad-positions. The control is the ad allocation from the baseline model. The organic content is not adjusted in any group. As performance metrics, we consider the average ad-CTR, average ad revenue, and average organics CTR across impressions in each group. Table~\ref{tab:externality} summarizes the lift estimates for these metrics. Looking at the first row, ads shuffling is seen to only barely affect organics, but to reduce the overall ads-CTR significantly. This suggests the order of the served ads matters. Looking at rows two and three, randomly selecting and randomly placing ads is not a good solution for ads-metrics as expected. Finally, because there is no change in the selection rule of organics compared to the baseline, the evidence seen of a significant impact on organics CTR due to changes in ads identity and location indicates an externality induced by ads.  These results motivate the importance of considering joint effects for ad allocation.

\begin{table}[h]
  \begin{threeparttable}
    \caption{Randomly allocating ads $-$ Dec 2-5, 2020 }
    \label{tab:externality}
    \centering
    \begin{tabular}{lllll}
    \toprule
                     & \multicolumn{3}{c}{Lift over the baseline model} \\
                      \cmidrule{2-4}
    Treat & \multicolumn{2}{c}{Advertisements} &   Organics & Sample\\
     group &       CTR &    Revenue &     CTR  & size\\
    \midrule
Shuffle         &    -1.44\%** &         -0.24\% &            0.03\% &  15.6M \\
$X=K_{ads}+1$        &    -1.90\%** &         -0.41\% &         1.32\%** &  15.6M \\
$X=K_{ads}+2$        &    -4.07\%** &      -3.01\%** &         1.41\%** &  15.7M \\
    \bottomrule
    \end{tabular}
    \begin{tablenotes}
      \small
      \item Sample size of control group: 31.3M
    \end{tablenotes}
    \end{threeparttable}
\end{table}

% \begin{table}[H]
%     \caption{Caption}
%     \label{tab:my_label}
%     \centering
%     \begin{tabular}{llllll}
%     \toprule
%     type & \multicolumn{4}{l}{ads} &   organic \\
%     var\_name &  bid\_price & charge\_price &      click &    revenue &     click \\
%     treat\_group &            &              &            &            &           \\
%     \midrule
%     top3        &  -0.12\%*** &     0.34\%*** &  -1.38\%*** &     -0.39\% &     0.03\% \\
%     top4        &  -0.97\%*** &     0.54\%*** &  -1.51\%*** &   -0.56\%** &  1.14\%*** \\
%     top5        &  -1.81\%*** &     0.84\%*** &  -3.44\%*** &  -2.74\%*** &  0.76\%*** \\
%     \bottomrule
%     \end{tabular}
%     \footnote{}
% \end{table}

\subsection{Virtual Bids}\label{subsec:vb_grid}
\begin{figure}[h]
  \centering
  \includegraphics[width=0.9\linewidth]{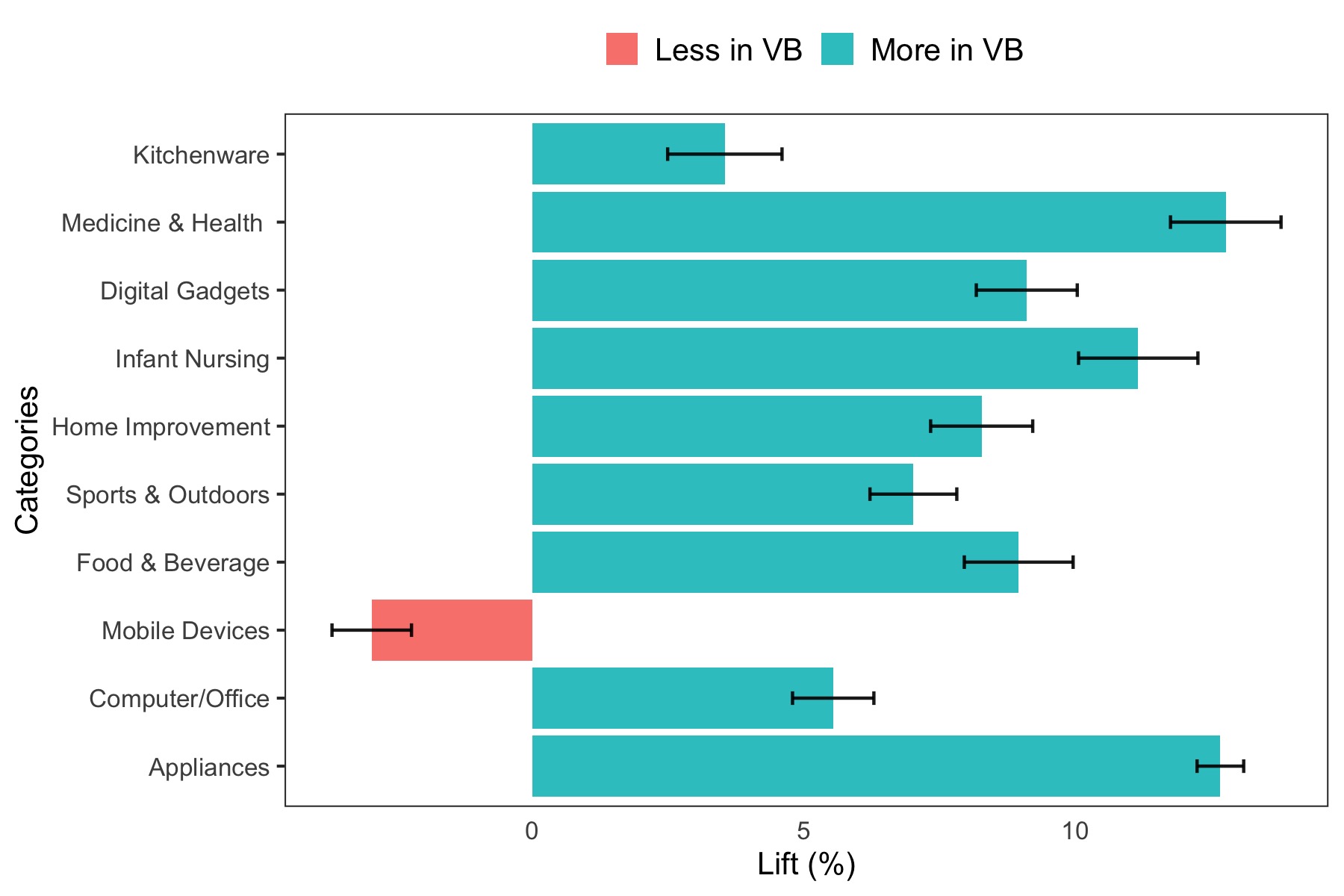}
  \caption{Distribution of ad revenue for optimal VB vs. baseline}
  \label{fig:vb_revenue_shift}
\end{figure}

\begin{figure}[h]
  \centering
  \includegraphics[width=0.8\linewidth]{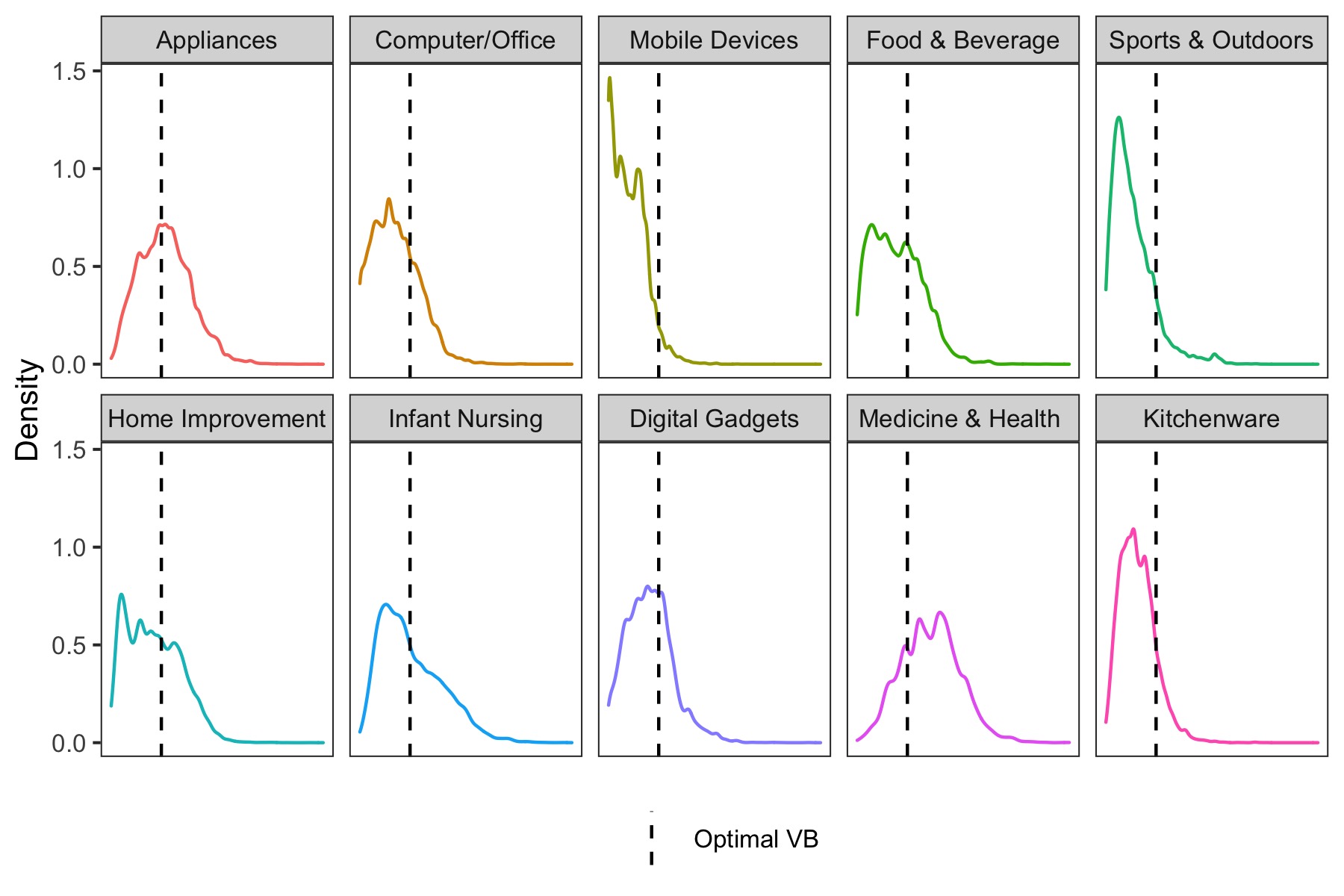}
  \caption{VB and Advertisers' Bid Distributions}
  \label{fig:dist_bid}
\end{figure}

\begin{table}[h]
  \begin{threeparttable}
    \caption{Grid around the optimal virtual bid $-$ Dec 15-17, 2020}
    \label{tab:vb}
    \centering
    \begin{tabular}{lllll}
    \toprule
                     & \multicolumn{3}{c}{Lift over the baseline model} \\
                      \cmidrule{2-4}
    Treat & \multicolumn{2}{c}{Advertisements} &   Organics & Sample\\
     group &       CTR &    Revenue &     CTR  & size\\
    \midrule
    $v^{a}$      &   1.06\%** &   6.60\%** &    -0.17\% &10.2M\\
    $v^{a} - 1$  &  -3.14\%** &  11.45\%** &  0.95\%** &10.2M \\
    $v^{a} + 1$  &   3.57\%** &   3.89\%** &    -0.12\% &10.2M\\
    \bottomrule
    \end{tabular}
    \begin{tablenotes}
      \small
      \item Sample size of control group: 20.5M
    \end{tablenotes}
    \end{threeparttable}
\end{table}
This experiment demonstrates the efficacy of the proposed approach compared to the baseline approach. Also, we demonstrate the proposed approach for choosing a virtual bid is able to find a tipping point for ad-CTR and ad revenue (i.e., improving both objectives to a point where improving one may worsen another). We consider three treatments corresponding to ad allocations that solve Eq. \ref{eq:virtual_bid} for three different values of $v^{a}$: (1) $v^{a}=$ the virtual bid chosen as discussed in Section~\ref{subsec:vb}; (2) $v^{a} - 1$; (3) $v^{a} + 1$. The control is the ad allocation from the baseline model. Looking at the first row in Table~\ref{tab:vb}, we see the proposed system is able to increase ad-CTR by $1.06\%$ and ad revenue by $6.60\%$ relative to the baseline system. In addition, the organics CTR is not statistically significantly different compared to the baseline for the treatment $v^{a}$. Rows two and three shows that $v^{a}$ is a tipping point and as we decrease or increase this virtual bid one of the objectives increases or decreases. Figure~\ref{fig:vb_revenue_shift} explores heterogeneity in ad revenue increases across product categories. Substantial heterogeneity is seen. The fact that the impact on ad revenue is different across categories (and is negative in one), suggests that fine-tuning of $v^{a}$ to specific categories could improve overall revenue even further. Figure~\ref{fig:dist_bid} compares $v^{a}$ vs. the CPC/bid distributions of advertisers across 10 selected categories. There are big disparities in the bid distribution across categories. This shows that even with full information of advertisers' bids/valuation, manual tuning of the virtual bids can be a difficult challenge for the platform. %Generally, the platform's WTP for clicks on ads chosen by our approach is similar to the median of several product categories, however, it may be higher for others. 

% \begin{table}[H]
%     \caption{Caption}
%     \label{tab:my_label}
%     \centering
%     \begin{tabular}{llllll}
%     \toprule
%     type & \multicolumn{4}{l}{ads} & organic \\
%     var\_name &  bid\_price & charge\_price &      click &   revenue &   click \\
%     treat\_group &            &              &            &           &         \\
%     \midrule
%     vb          &   4.45\%** &     5.50\%** &   1.01\%** &  4.78\%*** &  -0.11\% \\
%     vb\_minus1   &  12.89\%*** &    14.82\%*** &  -2.48\%*** &  8.03\%*** &   0.06\% \\
%     vb\_plus1    &  -1.05\%*** &    -0.55\%*** &   2.74\%*** &  2.42\%** &   0.02\% \\
%     \bottomrule
%     \end{tabular}
% \end{table}

\subsection{Impact on Diversity}

\begin{figure}[h]
  \centering
  \includegraphics[width=0.8\linewidth]{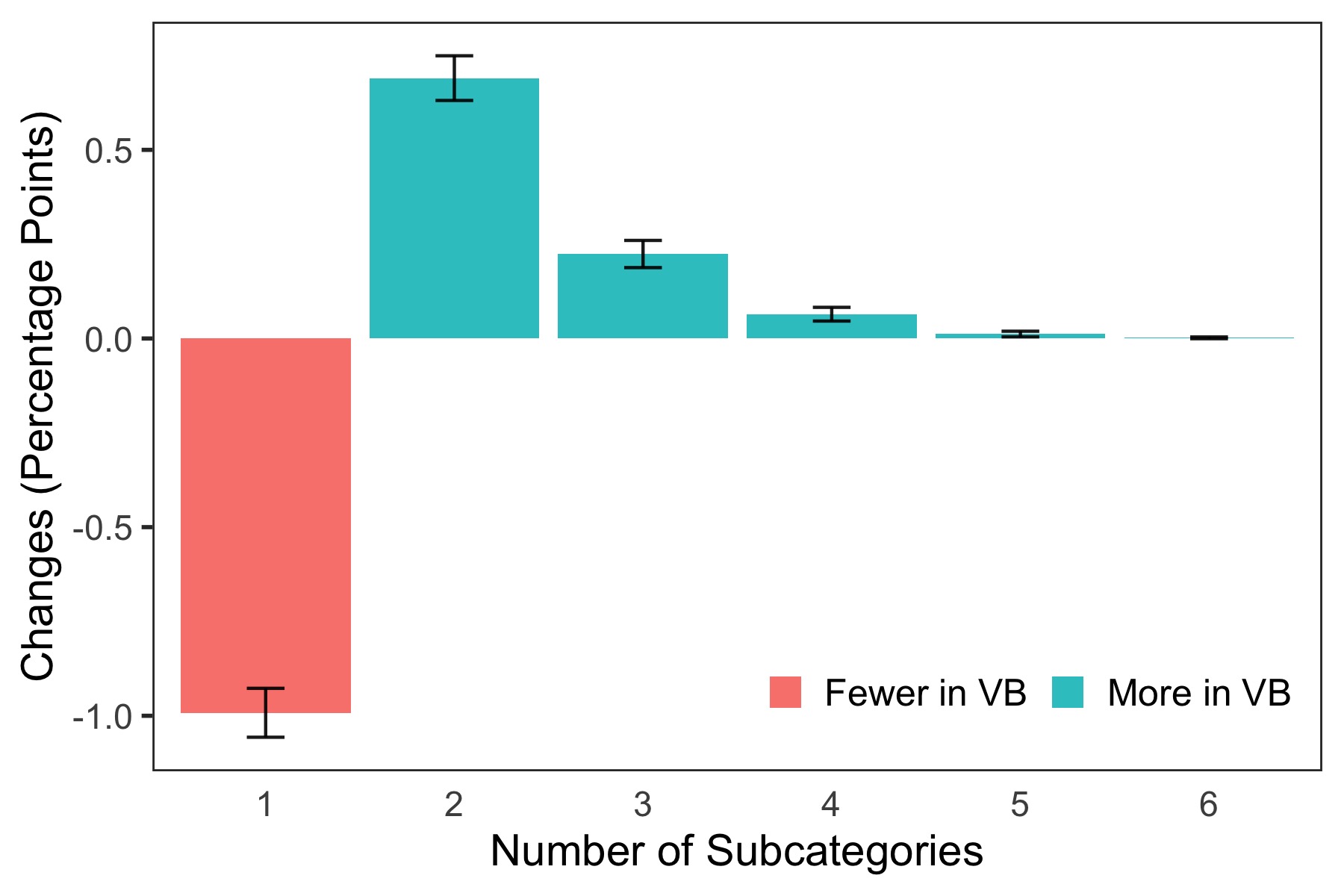}
  \caption{Distribution Changes of Number of Subcat. in the mixed tuple}
  \label{fig:diversity_hist}
\end{figure}

This experiment has one treatment and the control: the ad allocation with $v^{a}$ chosen as discussed in Section~\ref{subsec:vb} and the baseline. The goal of this experiment is to understand better possible sources of improvement under the new system. We demonstrate empirically that considering the joint effects explicitly induces the presentation of mixed tuples with greater diversity in terms of products. For the metrics, we consider: (1) More than 1 Subcat. in the mixed tuple meaning that at least one of the ads is of a product from a different subcategory than the product on whose page the ads are served; (2) Number of Subcat. in the mixed tuple meaning the number of different unique subcategories of products tied to the ads shown; and (3) Herfindahl-Index for the unique product subcategories of the ads shown (the index is a widely used measure of diversity, see \cite{rhoades1993herfindahl}). Looking at Table~\ref{tab:diversity}, we see the proposed approach is able to present more diverse mixed tuples than the baseline as measured in the three metrics. For the Herfindahl Index, a smaller value reflects more diversity. Notably, there is no change in diversity among organics as expected because the selection rule for organics is the same as in the baseline. Finally, Figure~\ref{fig:diversity_hist} shows this effect occurs not just at the mean: the distribution of number of subcategories shown to user under the proposed approach at the optimal virtual bid (termed ``VB'') is shifted to the right of the baseline. 

\begin{table}[h]
    \begin{threeparttable}
    \caption{Diversity for proposed system $-$ Jan 3-5, 2021}
    \label{tab:diversity}
    \centering
    \begin{tabular}{llll}
    \toprule
                     & \multicolumn{3}{c}{Lift over the baseline model} \\
     \cmidrule{2-4}
    Type & \makecell{More than 1 Subat. \\in the \emph{mixed tuple}} &
    \makecell{Number of Subcat. \\in the \emph{mixed tuple}} & Herfindahl \\
    \midrule
    ads     &  6.06\%** &  1.34\%** & -0.80\%**\\
    organic &    -0.16\% &    -0.02\% & 0.01\%  \\
    overall &    2.82\%** &    0.94\%** & -0.40\%** \\
    \bottomrule
    \end{tabular}
    \begin{tablenotes}
      \small
      \item Sample size of treatment group: 21.8M
      \item Sample size of control group: 21.8M
    \end{tablenotes}
    \end{threeparttable}
\end{table}

\subsection{Constant Virtual Bid (VB $=1$)}\label{subsec:VB=1}

This experiment features two treatments: ad allocations based on (1) $v^{a}$ chosen as  discussed in Section~\ref{subsec:vb}; (2) $v^{a} = 1$ reflecting an intuitive composite objective function maximizing the sum of ad revenue and ad-CTR. Ad allocation under the baseline remains as the control. The purpose of this experiment is to compare the data-driven proposed approach for choosing virtual bids vs. an intuitive virtual bid an analyst may consider. By analogy to the constrained version of the problem, another way to interpret the $v^{a} = 1$ treatment is it represents an ad-allocation under a corresponding ad-CTR threshold constraint that is not picked by principled tuning. We consider the same metrics discussed in previous experiments. Table~\ref{tab:cvb} indicates that the manually selected virtual bid $v^{a} = 1$ increases ad revenue but at the expense of ad-CTR relative to the baseline. In contrast, the data-driven selected virtual bid $v^{a}$ is able to find a balanced tipping point increasing ad-CTR by $2.05\%$ and ad revenue by $9.25\%$ relative to the baseline. It's worth noting this experiment was conducted separately from the experiment in Section \ref{subsec:vb_grid}, and row one serves as replication of the value of the new system. Being able to find a balanced tipping point both times showcases the reliability of our approach.

\begin{table}[h]
    \begin{threeparttable}
    \caption{Proposed system with VB $=1$ $-$ Dec 25-27, 2020}
    \label{tab:cvb}
    \centering
    \begin{tabular}{lllll}
    \toprule
                     & \multicolumn{3}{c}{Lift over the baseline model} \\
                     \cmidrule{2-4}
    Treat & \multicolumn{2}{c}{Advertisements} &   Organics & Sample\\
     group &       CTR &    Revenue &     CTR  & size\\
    \midrule
    $v^{a}$            &   2.05\%** &   9.25\%** &     0.32\% & 10.1M \\
    $v^{a}$=1         &  -4.79\%** &  13.57\%** &  1.19\%** & 10.1M\\
    \bottomrule
    \end{tabular}
    \begin{tablenotes}
      \small
      \item Sample size of control group: 20.2M
    \end{tablenotes}
    \end{threeparttable}
\end{table}

\subsection{Distribution Shift and Virtual Bids}\label{subsec:dist_shift}

\begin{figure}[h]
  \centering
  \includegraphics[width=0.8\linewidth]{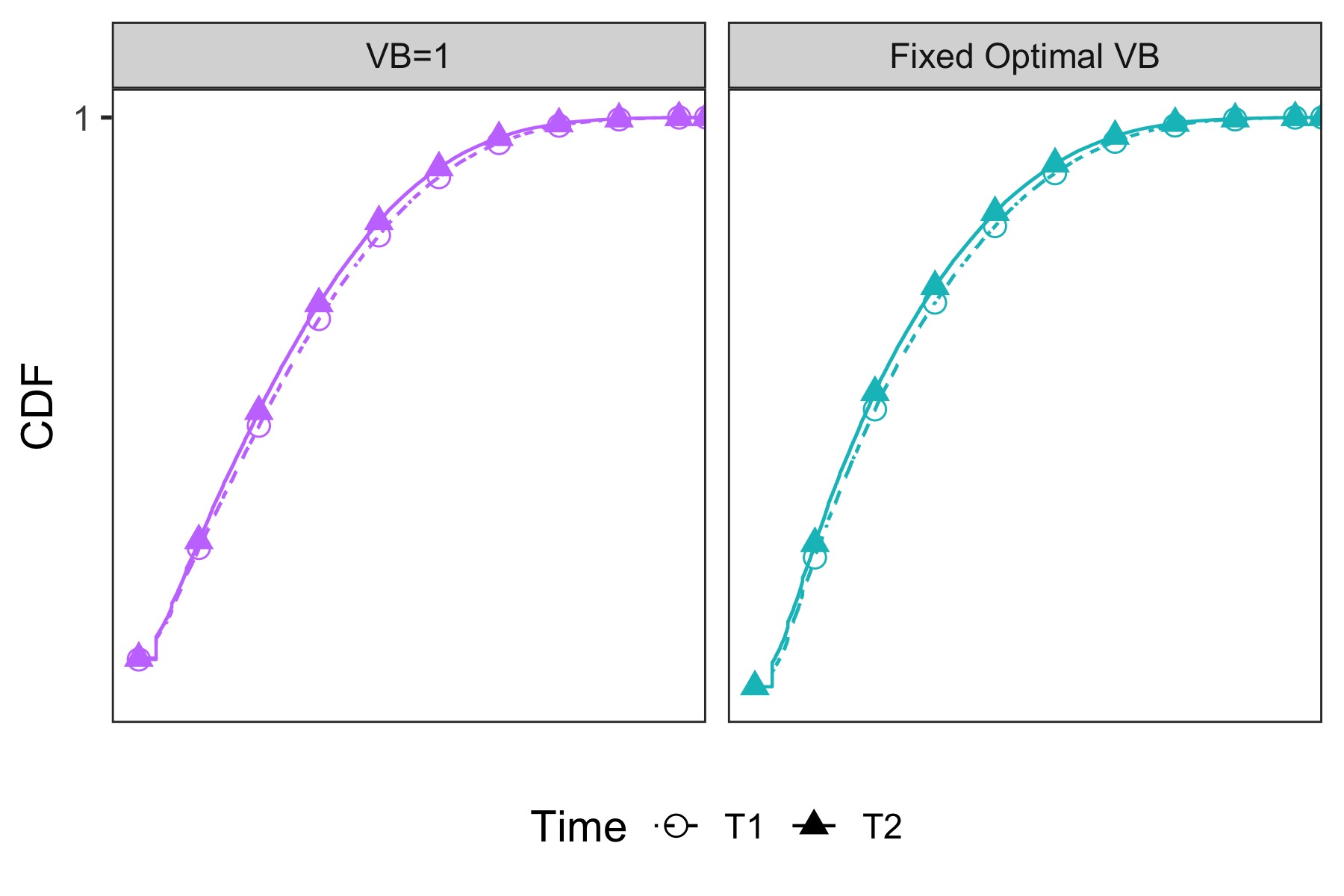}
  \caption{Distribution Shift in Revenue}
  \label{fig:dist_shift}
\end{figure}

The purpose of these experiments is to demonstrate the impact of distribution shifts in the environment, and how it may have an adverse impact on the metrics of interest if the virtual bids are not adjusted with proper frequency, or fixed manually to a static value. To implement this experiment, we repeat the two-treatment experiment described in section \ref{subsec:VB=1} above, two weeks later. For ease of exposition, we call the first the $T_1$-experiment and the second, implemented two weeks later, the $T_2-$experiment. The $T_2-$experiment is exactly the same as the $T_1-$experiment except that for its first treatment, we use the optimal value of $v^{a}$ obtained from the $T_1-$experiment. For its second treatment, we continue to use $v^{a}=1$. Then, for each treatment, we report in  Table~\ref{tab:ph}  the lift in various metrics between $T_2$ and $T_1$.
Looking at the table, we see that using the same virtual bid in $T_2$ as was optimal in $T_1$ leads to worse performance; also, holding the virtual bid constant at a fixed arbitrary value ($=1$) without any adjustment is bad for performance. Figure~\ref{fig:dist_shift} shows this occurs not just at the mean, documenting that the entire distribution of revenue is shifting between the two periods. Clearly, the virtual bids require adjustment in order to function properly, lending support to a data-driven approach such as the one proposed in this paper.

\begin{table}[h]
    \caption{Virtual Bids on two periods $-$ Dec 25-27, 2020 \& Jan 16-18, 2021}
    \label{tab:ph}
    \centering
    \begin{tabular}{llll}
    \toprule
     & \multicolumn{3}{c}{
     \makecell{Lift over respective T1 metrics\\
     of the same treatment}} \\
     \cmidrule{2-4}
    Treat & \multicolumn{2}{c}{Advertisements} &   Organics \\
     group &       CTR &    Revenue &     CTR  \\
    \midrule
     $v_{T_2}^{a}=v_{T_1}^{a}$ &   -0.33\% &  -8.55\%** &  -0.40\% \\
    $v_{T_2}^{a}$ = 1 &   -1.61\%** &  -9.33\%** &  -0.64\%* \\
\bottomrule
    \end{tabular}
\end{table}

% \begin{table}[H]
%     \caption{Caption}
%     \label{tab:my_label}
%     \centering
%     \begin{tabular}{llllll}
%     \toprule
%     type & \multicolumn{4}{l}{ads} &   organic \\
%     var\_name &  bid\_price & charge\_price &      click &    revenue &     click \\
%     treat\_group &            &              &            &            &           \\
%     \midrule
%     vb          &   5.64\%*** &     6.83\%*** &   1.40\%*** &   6.42\%*** &  0.65\%*** \\
%     vb=1        &  15.87\%*** &    18.37\%*** &  -3.29\%*** &  10.47\%*** &  1.16\%*** \\
%     \bottomrule
%     \end{tabular}
% \end{table}
\section{Related Work}
\label{sec:relwork}
Our work intersects with the multiobjective optimization of recommendation systems for the allocation of contextual ad content. In the contextual ad-recommendation literature, the main thrust is on developing systems for serving recommendations considering the additional information available in the environment, e.g., a user interacting with a related item, which provides the context for recommendations. \cite{barbieri2014, agarwal2016} provide recent treatises on this literature, and \cite{zhang2019deep, zhao2019deep} provides a review from the perspective of deep learning systems using supervised learning and/or reinforcement learning approaches. Within this stream, our work fits into feature-based deep supervised learning systems for item recommendations. A novelty of our approach is that we consider joint effects in the CTR prediction. In our approach, context is not only the product visited by a user but also the recommended ads and organics displayed jointly to the user. Thus, we do not make the typical assumption of conditional independence of the CTRs of the ads and organics displayed jointly given the context of the product featured.  In addition, we show how we can integrate this prediction system in a congruent way into an allocation scheme that respects multiple, potentially competing objectives of interest, an open issue for recommendation systems in general (see \cite{adomavicius2011multi}).

Within the multiobjective optimization problem of recommendation systems, our work is closest to the stream of \emph{constrained optimization} approaches proposed in \cite{agarwal2011click, agarwal2012personalized, yan2020}. \cite{agarwal2011click} proposed a linear constrained optimization problem to optimize the engagement time spent of users on an article's landing page subject to CTR constraints. The minimum desired CTRs are a fraction of the optimal CTRs obtained by maximizing an unconstrained maximum CTR problem. Thus, this fraction is effectively a hyperparameter that requires tuning.  \cite{agarwal2012personalized} improves on \cite{agarwal2011click} by allowing the linear constrained optimization problem of \cite{agarwal2011click} to be personalized, and uses Lagrangian Duality to construct a scalable solution algorithm. \cite{yan2020} also formulate a constrained optimization and propose a scalable solution algorithm where the Lagrangian Duals are taken as given hyperparameters. For optimal performance, these duals have to be tuned. While there are many suggestions, developing a structured approach to tuning such hyperparameters remain an open question for the literature.

In contrast, our work presents an \emph{unconstrained optimization} formulation of the problem. The advantage of this relative to constrained approaches was discussed earlier in section \ref{sec:compare-constrained-approaches}. More broadly, a novel contribution of this study is to present a principled approach to tuning. We propose an approach using a bilevel optimization problem to find the optimal values for the hyperparameters of our problem (the virtual bids), which to the best of our knowledge has not been proposed in the ads recommendation literature. We show that tuning the virtual bids appropriately is critical for optimal performance. Furthermore, our work is easily extendable to a setting of optimal allocation of both ads and organics with multiple competing objectives similar to \cite{yan2020}.

Theoretical issues regarding locating pareto optimal solutions to multiobjective problems have been presented in the literature on compromise optimization; see, for e.g., \cite{miettinen}. Our contribution here is more applied, in applying these ideas to the specific problem of ad-recommendation in e-commerce, considering externalities, and showing empirical evidence of its value in the context of a large-scale, deployed application.

%\cite{zhao2020} DRL for multiple objectives

Lastly, our work forms the allocation scheme of an Auction Design problem \cite{nisan2007intro, tim2016}. The other important part of an Auction Design problem is the payments scheme. Our formulation links our allocation scheme to a VCG payment scheme which is attractive for contextual advertising with externalities, see \cite{lahaie2007, varian2014vcg}. In computing VCG payments, the objective in Eq. \ref{eq:virtual_bid} also serves as the VCG$-$auction payoff, enabling linking the allocation and payment parts in an internally-consistent way. While our proposed allocation scheme may be used with GSP auctions, the \emph{externalities} present would not be properly priced under the GSP mechanism unlike the VCG mechanism (see \cite{gomes2009externalities, roughgarden2012externalities}). In the version currently deployed, the payment scheme continues to use a GSP scheme because moving from a GSP auction system to a VCG auction system involves complex business and engineering considerations requiring further evaluation. The advantage of setting up the ad allocation in the manner we presented here is it accommodates a seamless change to a VCG auction system, pending such an evaluation.

\section{Conclusion}\label{sec:conclusion}
We propose a system for ads allocation in the presence of organic content on product detail pages in e-commerce, and report on the system's deployment on \emph{JD.COM}'s mobile app. Our system has 3 key features: (1) Optimization of multiple competing business objectives through a virtual bids approach; (2) Modeling of users' click behavior considering explicitly joint effects through a deep learning approach; 3) Consideration of externalities in the allocation of ads and direct compatibility with the VCG auction's payment scheme, which accommodates such externalities. We demonstrate empirically the presence of motivating joint effects in our application. Also, we show that considering these effects explicitly has benefits such as increased diversity in the ads presented. Also, we demonstrate that the proposed data-driven approach for choosing virtual bids is able find tipping points for the objectives. Lastly, we highlight the importance of a data-driven approach to deal with distribution shifts in the data and emphasize the importance of properly updating these bids for good performance. 

\bibliography{vcg-rec}
\pagebreak

\appendix

\begin{comment}
\section{Appendix: Utopia Point}\label{app:utopia-plot}
This appendix shows graphical intuition for the way the virtual bids in Sec.~\ref{subsec:vb} are located by finding a tuple on the possibility frontier of ad-CTR and ad revenue that is closest to the \emph{Utopia Point}.
\begin{figure}[h]
  \centering
  \includegraphics[width=0.85\linewidth]{stylized_plot.jpeg}
  \caption{Stylized Plot of Bi-dimensional Possibility Frontier and Utopia Point}
  \label{fig:stylized_plot}
\end{figure}  
\end{comment}
\section{Appendix: Golden Search and Simultaneous Perturbation Stochastic Approximation (SPSA)}\label{app:optim-algos}

This appendix summarizes the \emph{Golden Search} and the \emph{Simultaneous Perturbation Stochastic Approximation (SPSA)} methods for the optimization of the bilevel level optimization problem described in Section~\ref{subsec:vb}.

\subsection{Golden Search}

The Golden Search is a \emph{line search} algorithm for finding the minimum of a one dimensional function in an interval. The method proceeds by narrowing the feasible space to search over through specifying successive subintervals to search over where the length of these subintervals is based on the \emph{Golden section ratio} (i.e. $\phi = \frac{1 + \sqrt{5}}{2}$), see \cite{luenberger2010}. Algorithm~\ref{alg:gs} presents pseudo code for its implementation.

\begin{algorithm}[h]
\caption{Golden Search}
\label{alg:gs}
\KwIn{$F(\cdot)$: Function of the upper level specified in Eq.~\ref{eq:bilevelopt}; $[a,b]$: Interval of feasible values; $K$ maximum number of iterations; $tol$ tolerance.}
\KwOut{$v^{a}$: Optimal Virtual Bid}
Initialize $\phi = \frac{1 + \sqrt{5}}{2}$; $d = b - a$; $l = a + \frac{d}{\phi}$; $u = b - \frac{d}{\phi}$; $v^{a} = a$ \\
\For{$k \gets 1$ \textbf{to} $K$} {
  \uIf{$F(u) <  F(l)$} {
    $b = u$
  }
  \Else {
   $a = l$
  }
  $d = b - a$; $l = a + \frac{d}{\phi}$; $u = b - \frac{d}{\phi}$
  
  \If{$|b-a| < tol$} {
  $v^{a} = a$ \\
  \textbf{break}
  }
  
}
\KwRet{$v^{a}$}

\end{algorithm}

\subsection{Simultaneous Perturbation Stochastic Approximation (SPSA)}

SPSA is a \emph{stochastic approximation} algorithm where the gradient of the objective function is estimated using its own measurements. SPSA only requires $2$ function evaluations per iteration regardless of the dimension of the problem, and thus this makes it suitable for large-scale optimization. In addition, the noisy gradient estimator of SPSA (i.e. \emph{Simultaneous Perturbation} $\vec{g}$) provides it with resistance to being stuck in local optima \cite{spall1998}. Algorithm~\ref{alg:spsa} presents pseudo code for its implementation.

\begin{algorithm}[H]
\caption{SPSA}
\label{alg:spsa}
\KwIn{$F(\cdot)$: Function of size $p$ for the upper level specified in Eq.~\ref{eq:bilevelopt}; $K$ maximum number of iterations; $\alpha$, $\gamma$, $A$, $a$, $c$: hyper-parameters of SPSA, see \cite{spall1998} for instructions on tuning.}
\KwOut{$\vec{v}$: Optimal Virtual Bids vector of size $p$}
Initialize $a_{k} = 0$; $c_{k} = 0$; $\vec{g}_k = \vec{0}$; $\vec{\theta}_{k} = \vec{0}$
\For{$k \gets 1$ \textbf{to} $K$} {
$a_{k} = \frac{a}{(k + A)^{\alpha}}$ ;
$c_{k} = \frac{c}{(k)^{\gamma}}$ \\
Sample $\vec{\Delta}_{k}$ of size $p$ from $2(Bernoulli(0.5)) - 1$ \\
$\vec{\theta}_{plus} = \vec{\theta}_{k} + c_{k}\vec{\Delta}_{k}$  \\
$\vec{\theta}_{minus} = \vec{\theta}_{k} - c_{k}\vec{\Delta}_{k}$  \\
$y_{plus} = F(\vec{\theta}_{plus})$ ; $y_{minus} = F(\vec{\theta}_{minus})$ \\ 
$\vec{g}_{k} = \frac{y_{plus} - y_{minus}}{2 c_{k}\vec{\Delta}_{k}}$  (elementwise operation) \\
$\vec{\theta}_{k} = \vec{\theta}_{k} - a_{k}\vec{g}_{k}$
}
$\vec{v} = \vec{\theta}_{k}$ \\
\KwRet{$\vec{v}$}

\end{algorithm}

\end{document}